\documentclass[10pt, a4paper, onecolumn]{article}

\usepackage[numbers]{natbib}

\usepackage[T1]{fontenc}
\usepackage{lmodern}

\usepackage{natbib}
\usepackage[utf8]{inputenc} 
\usepackage{booktabs}       
\usepackage{amsfonts}       
\usepackage{nicefrac}       
\usepackage{microtype}      
\usepackage{xcolor}         
\usepackage{graphicx}
\usepackage{amsmath,amssymb,amsfonts}
\usepackage{bm}

\usepackage{url}
\usepackage{hyperref}

\usepackage{algorithm}
\usepackage{algorithmicx}
\usepackage{algpseudocode}

\usepackage{array,multirow,graphicx}

\usepackage{geometry}
\newgeometry{vmargin={15mm}, hmargin={12mm,17mm}} 

\title{GroupGazer: A Tool to Compute the Gaze per Participant in Groups with integrated Calibration to Map the Gaze Online to a Screen or Beamer Projection}

\author{
	Wolfgang Fuhl, Daniel Weber, Shahram Eivazi\\
	Department of Human Computer Interaction\\
	University Tübingen\\
	Tübingen, 72076 \\
	\texttt{wolfgang.fuhl@uni-tuebingen.de} \\
	\texttt{daniel.weber@uni-tuebingen.de} \\
	\texttt{shahram.eivazi@mnf.uni-tuebingen.de} \\
}

\begin{document}
	
	\maketitle
	
	\begin{abstract}
		In this paper we present GroupGaze. It is a tool that can be used to calculate the gaze direction and the gaze position of whole groups. GroupGazer calculates the gaze direction of every single person in the image and allows to map these gaze vectors to a projection like a projector. In addition to the person-specific gaze direction, the person affiliation of each gaze vector is stored based on the position in the image. Also, it is possible to save the group attention after a calibration. The software is free to use and requires a simple webcam as well as an NVIDIA GPU and the operating system Windows or Linux. \\
		\url{https://es-cloud.cs.uni-tuebingen.de/d/8e2ab8c3fdd444e1a135/?p=%2FGroupGazer&mode=list}.
	\end{abstract}

	\section{Introduction}
	\begin{figure}
		\centering
		\includegraphics[width=\textwidth]{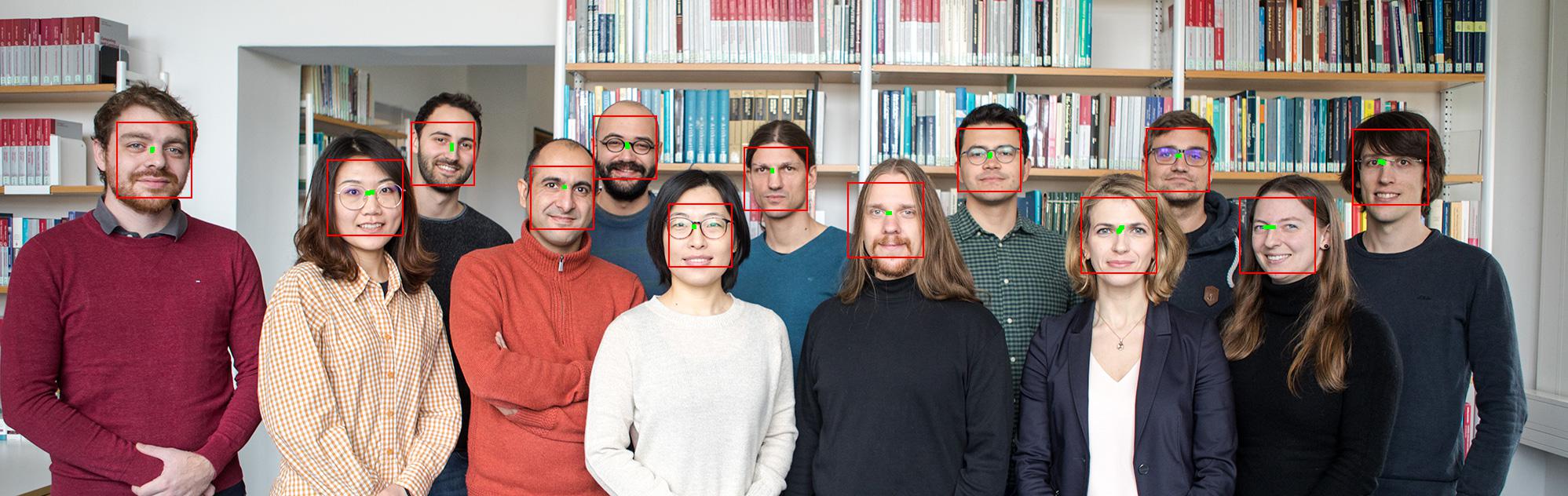}
		\caption{Detection results of the proposed tool GroupGazer on a group photo.}
		\label{fig:teaser}
	\end{figure}

Eye tracking is an important input modality and information source in the modern world~\cite{cognolato2018head,AGAS2018,WF042019,ROIGA2018,ASAOIB2015}. The signal itself is usually computed by first detecting the pupil~\cite{WTCKWE092015,WTTE032016,062016,CORR2017FuhlW2,CORR2016FuhlW,WDTE092016} and additional eye features~\cite{ICML2021DSISMAR,ICCVW2019FuhlW,CAIP2019FuhlW,ICCVW2018FuhlW}. Based on the gaze signal it is possible to extract further information like the eye movement types~\cite{FCDGR2020FUHL,fuhl2018simarxiv,ICMIW2019FuhlW1,ICMIW2019FuhlW2,EPIC2018FuhlW,MEMD2021FUHL}. In the field of human-machine interaction~\cite{UMUAI2020FUHL}, the gaze signal is used and further researched for interaction with robots~\cite{willemse2019natural} but also other technical devices~\cite{wanluk2016smart,C2019,FFAO2019,NNVALID2020FUHL}. This involves not only simple control but also collaboration in which a human communicates complex behavior to a robot or system~\cite{palinko2016robot}. Interaction with the eyes based on pupil movements~\cite{WTCDAHKSE122016,WTCDOWE052017,WDTTWE062018,VECETRA2020,ETRA2018FuhlW,ETRA2021PUPILNN} is also an interesting source of information in the field of computer games~\cite{alkan2007studying}. Eye interaction is many times faster than mouse interaction, which could revolutionize the professional computer gaming field~\cite{jonsson2005if}. In the field of virtual reality, gaze information can be used to render only small areas of the scene in high resolution, leading to a significant reduction in the resources consumption of the devices~\cite{meng2018kernel}. Another important area in which the gaze signal plays an important role is driver observation. Here it is necessary to assess whether the driver is able to control the vehicle or is too tired in the case of autonomous driving to take over the vehicle~\cite{zandi2019non}. Of course, this also applies to car rental companies, for which it is important to know whether the driver is, for example, intoxicated or an unsafe driver~\cite{maurage2020eye,WTDTWE092016,WTDTE022017,WTE032017,NNETRA2020,CORR2017FuhlW1}. In the field of medicine, research is also being conducted into methods of self-diagnosis~\cite{clark2019potential}. This involves, for example, the early detection of Alzheimer's disease~\cite{crawford2015disengagement}, strokes~\cite{matsumoto2011neurologists}, as well as eye defects~\cite{eide2019detecting} or autism~\cite{boraston2007application}. In the field of safety, the eye signal also gains increasingly more interest~\cite{katsini2020role,RLDIFFPRIV2020FUHLICANN}. This is due to the fact that personal behavior is reflected in the gaze signal, which can be used to identify the person~\cite{GMS2021FUHL,RLDIFFPRIV2020FUHLICANN}. Other information contained in the eye is the cognitive load based on pupil dilation~\cite{chauliac2020all}, attention~\cite{chita2016social}, procedural strategies~\cite{jenke2021using} and many others. A relatively new area in which the eye tracking signal is used is behavioral research~\cite{yang2021webcam,das2018social}. Here it is on the one hand about extracting expert knowledge from the eye signal and passing this knowledge to trainees~\cite{manning2003eye,hoghooghi2020novice}. This concerns, all areas in which the training is only possible with expensive tools and training devices~\cite{vijayan2018eye}. In the area of medicine the main interest is to distill the expert knowledge better~\cite{quen2021medical,manning2003eye}. Another area of behavioral research which is also the subject of this thesis is group behavior~\cite{hwang2020eye,reichenberger2020gaze,kredel2017eye}. Here there is research in the area of teaching~\cite{korbach2020should,schneider2008validation,jarodzka2020eye} but also in dynamic environments like sports~\cite{oldham2021association,du2009sport,reneker2020virtual}.

The current problems in the field of behavioral research for groups, is that there is no freely available software for this. Therefore, research groups have to resort to expensive solutions such as multiple worn eye trackers. This creates further issues like the assignment of the important areas between the different scene cameras. One way around this is to use virtual reality together with eye tracking. However, this also changes the behavior of the test persons and cannot be carried out over longer periods of time with regard to motion sickness. Alternatively to worn eye trackers, there is also the possibility to use external cameras. In this case, the researchers have to implement their technical solutions independently, which often leads to dependencies on other working groups and is also an expensive undertaking due to the image processing cameras which are usually used. 

In this paper, we present software that allows anyone to use a simple webcam for gaze estimation of groups and calibration each subject in parallel. By doing so, we hope to enable anyone to conduct behavioral group research. Our contributions to the state of the art are:
\begin{enumerate}
	\item A tool to record the gaze of groups and calibrate each individual in parallel.
	\item The tool has no specialized hardware requirements and only needs an NVIDIA GPU with at least 4 GB memory (We used a 1050 ti with 4 GB).
	\item Stores the gaze per person as well as the average gaze location of the group.
\end{enumerate}

\section{Related work}
Since our software is the combination of several research fields, we have divided the related work into three categories. The first category is face recognition, the second category is appearance based gaze estimation, and the third category is gaze based group behavior research.

\subsection{Face detection}
Face recognition in arbitrary environments is still a very challenging field of research. Here, an arbitrarily large image is given, and all faces must be detected. This often involves occlusions, different head positions, changes in lighting conditions, and of course the faces in the image have different resolutions. The first very successful approach was presented by Viola and Jones~\cite{142aaa}. This is based on hair features and trained using AdaBoost. The next major step was achieved with deformable part models (DPM)~\cite{158aaa}. Compared to feature-based approaches, DPM is much more robust but requires significantly more computational effort. With the advent of deep neural networks, however, the state of the art was again significantly improved~\cite{6aaa,63aaa,79aaa,182aaa}. The first extension of neural networks was the combination of face detection with face matching~\cite{172aaa,176aaa}. Current methods for face detection follow two directions. The first direction is the multistage approach, which is based on a region proposal neural network followed by validation of the proposed faces. The most notable representatives of this direction of development are RCNN~\cite{39aaa}, almost RCNN~\cite{38aaa}, and faster RCNN~\cite{119aaa}. The second direction of development is direct methods such as single shot multibox detector (SSD)~\cite{85aaa} or you only look once (YOLO)~\cite{117aaa}. For YOLO, there are already multiple versions, which consume even less resources at approximately the same detection rate. The advantage of the direct approaches, is the faster execution and the smaller resource consumption. The multilayer methods, on the other hand, provide a better detection rate and fewer misclassifications. Further ressource consumption approaches have also been proposed in the literature~\cite{AAAIFuhlW,NNPOOL2020FUHL,RINGRAD2020FUHL}.

\subsection{Appearance based gaze estimation}
Here, the entire facial image or eye area of a person is used to directly determine the gaze vector via a neural network. The first work in this area is from 1994~\cite{27abc} and was extended in \cite{15abc} by linear projection functions. These methods require very expensive calibration, since the neural network was trained for each person individually with many training examples. The first extensions to reduce the effort in calibration were a Gaussian process regression~\cite{16abc}, saliency maps~\cite{28abc}, and optimal selection using a linear regression~\cite{13abc}.  While all of these methods advanced the state of the art, the appearance based approach still had many limitations, such as a fixed head position and per-person calibration. With deep neural networks and the advent of big data, this has changed significantly. In \cite{26abc} the first successful approach was presented, which realized appearance based gaze estimation with deep neural networks. The first extensions used, in addition to eye images, the subjects' faces, which resulted in a significant improvement~\cite{35abc,kellnhofer2019gaze360}. For extreme head positions and strongly deviating gaze angles to the head orientation, an asymmetric regression was presented~\cite{37abc}.

\subsection{Gaze based group behavior research}
In this section, we would like to mention and briefly explain only some works from this area, since our software is made for this purpose but does not perform a behavior research study. 

The first area in behavioral research which can also be applied to groups is mind wandering~\cite{1xxx,hutt2019automated}. Mind wandering is a shift in attention to task-unrelated thoughts. This is an interesting effect for teaching since it negatively influences the learning performance of students~\cite{50xxx,56xxx,hutt2019automated}. Mind wandering itself is a special form of disengagement and has to be separated from boredom or off-task behaviors~\cite{32xxx,36xxx,hutt2019automated}. Another interesting social behavior is gaze following~\cite{aung2018they}. This gaze following is a form of communication and socializing. In some scenarios it has to be done only for a single person~\cite{15yyy,11yyy} but in modern research entire scenes with multiple persons are evaluated and analyzed~\cite{aung2018they,24yyy,21yyy,28yyy,27yyy}. Nowadays, psychologists use behavior observation methods in classrooms as well as direct behavior ratings~\cite{woolverton2021exploration}. While both methods are valid and also used by teachers themselves, they are limited in effectiveness due to the attentional limits of the human observers as well as their induced biases~\cite{mcintyre2018scanpath}. Modern research focuses on establishing intelligent classroom technologies with eye tracking and voice recording~\cite{woolverton2021exploration,mcintyre2018scanpath}. Those methods have their limitations due to the data security but deliver more insights and allow reducing the induced bias by humans~\cite{woolverton2021exploration,mcintyre2018scanpath,mcparland2021investigating}.

\section{Method}
\begin{figure*}
	\centering
	\includegraphics[width=0.9\textwidth]{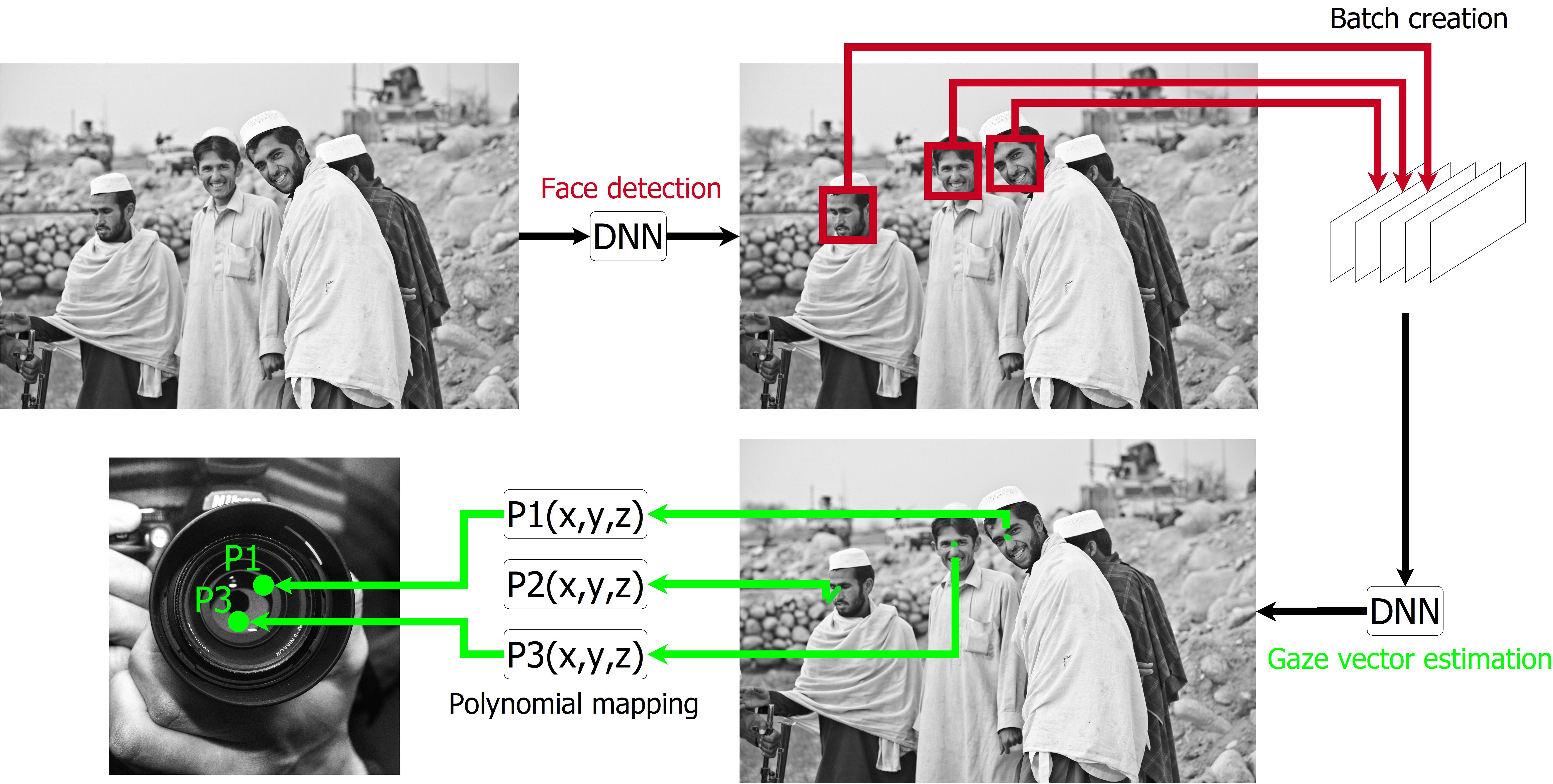}
	\caption{The workflow of our approach. We first detect the faces and compute the gaze vector using an appearance based approach. Each person is calibrated using mouse clicks on a projection in parallel. The fitted polynomials are afterwards used to map the gaze to the projection.}
	\label{fig:workflow}
\end{figure*}

\begin{table}[htb]
	\caption{Shows the architecture of our face detection deep neural network. The architecture is copied from dlib~\cite{king2009dlib} and uses the max margin~\cite{king2015max} training procedure. We modified the model in terms of tensor normalization~\cite{2021TNandFDT} and gradient centralization~\cite{NORM2020FUHLNEU} as well as convolution size and depth.}
	\label{tbl:networkarchDetect}
	\centering
	\begin{tabular}{ll}
		\toprule
		Level & Gaze estimator \\
		\midrule
		Input & RGB image any resolution\\
		1 & Pyramid layer with six stages \\
		2 & $5 \times 5$~Conv, depth 8, $2 \times 2$~down scaling, BN, ReLu with tensor normalization \\
		3 & $3 \times 3$~Conv, depth 8, $2 \times 2$~down scaling, BN, ReLu with tensor normalization \\
		4 & $3 \times 3$~Conv, depth 8, $2 \times 2$~down scaling, BN, ReLu with tensor normalization \\
		5 & $5 \times 5$~Conv, depth 16, BN, ReLu with tensor normalization \\
		6 & $3 \times 3$~Conv, depth 16, BN, ReLu with tensor normalization \\
		7 & $3 \times 3$~Conv, depth 16, BN, ReLu with tensor normalization \\
		8 & $7 \times 7$~Conv, depth 1 \\
		\bottomrule
	\end{tabular}
\end{table}

\begin{table}[htb]
	\caption{Shows the architecture of our gaze estimation deep neural network. It has the structure of a ResNet-34~\cite{he2016deep} and uses the leaky maximum propagation blocks~\cite{fuhl2021maximum}, tensor normalization~\cite{2021TNandFDT}, as well as the weight and gradient centralization~\cite{NORM2020FUHLNEU}.}
	\label{tbl:networkarchGaze}
	\centering
	\begin{tabular}{ll}
		\toprule
		Level & Gaze estimator \\
		\midrule
		Input & Gray scale image $100 \times 100$ \\
		1 & $5 \times 5$~Conv, depth 32 \\
		2 & ReLu with tensor normalization \\
		3 & $2 \times 2$~Max pooling \\
		4 & 3 Maxium connection blocks, $2 \times 2$~down scaling, $3 \times 3$~Conv, depth 64, BN  \\
		5 & ReLu with tensor normalization \\
		6 & 3 Maxium connection blocks, $2 \times 2$~down scaling, $3 \times 3$~Conv, depth 128, BN  \\
		7 & ReLu with tensor normalization \\
		8 & 3 Maxium connection blocks, $2 \times 2$~down scaling, $3 \times 3$~Conv, depth 256, BN  \\
		9 & ReLu with tensor normalization \\
		10 & Fully connected, 512 outputs \\
		11 & ReLu \\
		12 & Fully connected, 7 outputs (3 and 7 are the accuracy of the estimation~\cite{ICMV2019FuhlW})\\
		\bottomrule
	\end{tabular}
\end{table}

Figure~\ref{fig:workflow} shows the workflow of our approach. GroupGazer first opens a video stream on an available camera. Afterwards, all faces in the image are detected. If not all desired faces are detected, GroupGazer offers an upscaling factor, which can be set by the user. This upscaling factor resizes the input image to allow the face detection to detect even very small faces in the image. After the face detection, all detected faces are extracted from the image and resized to $100 \times 100$ pixels in a gray scale image. These images are grouped together to form a batch which is given to the gaze vector estimation DNN. The batch size can also be set by the user. This fixed batch size allows GroupGazer to have a static runtime and if there are fewer faces in the image, the rest of the batch is filled with black images. GroupGazer can be used with a 1050 ti graphics card for up to 40 faces in real time, which is also dependent on the input resolution to the face detection DNN. For newer GPUs more faces can be set by the user as well as larger input image resolutions for the face detection.
The gaze estimation DNN processes the entire batch and computes a starting position (First two values), an accuracy of the starting position (Third value), the gaze vector (Forth to sixth value), as well as an accuracy of the gaze vector (Seventh value). With this information, each face has a gaze vector and an estimated accuracy. 
With the gaze vector and the starting position, a polynomial is used to map the gaze vector to a projection or monitor. The degree of the polynomial can be specified by the user, and the calibration procedure works as follows. The teacher or adviser tells the students to look at his mouse cursor position. On a left mouse click, all gaze vectors which are seen as valid and accurate are stored together with the click location. This is repeated multiple times. Afterwards, for each user, the polynomial is fitted in the least squares sense. With those polynomials, the mapping and therefore the gaze location is computed for each user.
The reidentification of users is done by the smallest euclidean distance to the last detections, and the new position is not allowed to leaf the last face detection bounding box. This is a simple procedure but saves a lot of computational resources since no additional network has to be used. In addition, it is much more robust since fine-tuning a Network online usually needs multiple examples to deliver reliable results, even if we use the hypersphere approach~\cite{xie2019deep} or siam networks~\cite{abdelpakey2019dp}.

The used model architectures can be seen in Table~\ref{tbl:networkarchDetect} and \ref{tbl:networkarchGaze}. Our face detection model is similar to the model from dlib~\cite{king2009dlib} we only made some slight changes which improve the accuracy of the model and only impact the runtime slightly. For gaze estimation we used the architecture of a ResNet-34~\cite{he2016deep} since it has a good accuracy and is resource saving in contrast to the other networks. We modified the ResNet-34 architecture only by adding some novel normalization~\cite{2021TNandFDT,NORM2020FUHLNEU}, the landmark validation loss~\cite{ICMV2019FuhlW}, as well as leaky maximum propagations instead of the residual connections~\cite{fuhl2021maximum}. 


\section{Evaluation}

\textbf{Gaze360}~\cite{kellnhofer2019gaze360} is a huge data set with 3D gaze annotations recorded using multiple cameras covering 360 degree. The recordings were conducted indoor and outdoor with 238 subjects. The dataset contains large head variations as well as distances of the subjects to the camera. We only used approximately 80,000 images of this data set since the data set contains also human heads from behind as well as some partially covered heads which we removed from our data for training and evaluation. The train and test split was done by randomly selecting 20\% for testing and 80\% for training.

\textbf{DLIB}~\cite{king2009dlib} data set contains images of various resolutions. Each image can have multiple faces which are annotated with bounding boxes. In total the data set contains 7213 images and 11480 annotated faces. We made a random 50\% to 50\% split and used the first half for training and the second half for evaluation (One image more for training due to the uneven number). The images in the dataset are taken from other public data sets and annotated by the authors of \cite{king2009dlib}.

\begin{table}[htb]
	\caption{Face detection results on DLIB data set~\cite{king2009dlib} with percition and recall. We compare our model to other approaches in therms of detection percentage as well as runtime in milliseconds (ms) for one hundred images in average. OoM means out of memory exception.}
	\label{tbl:evalFDDB}
	\centering
	\begin{tabular}{lcccc}
		\toprule
		Method & Percision & Recall & \multicolumn{2}{c}{Runtime GPU (ms)} \\
		& & & $1920 \times 1280$ & $300 \times 300$ \\
		\midrule
		Proposed & 0,99 & 0,89 & 67 & 3 \\
		dlib std. arch.~\cite{king2009dlib} & 0,99 & 0,88 & 175 & 8 \\
		ResNet-34 \& Faster-RCNN~\cite{119aaa} & 0,99 & 0,91 & OoM & 22 \& 1\\
		Yolov5s~\cite{117aaa} & 0,99 & 0,89 & OoM & 10\\
		\bottomrule
	\end{tabular}
\end{table}

\begin{table}[htb]
	\caption{Appearance based gaze estimation results on the Gaze360~\cite{kellnhofer2019gaze360} dataset. We compared our model to other approaches and evaluated the gaze start estimation in average euclidean distance in pixel as well as the gaze vector estimation in degree. Time is measured for one face image as average over one thousand.}
	\label{tbl:evalGZ}
	\centering
	\begin{tabular}{lccc}
		\toprule
		Method & Gaze start   & Gaze vector & Runtime GPU (ms) \\
		\midrule
		Proposed & 0,6 & 0,2 &  3 \\
		ResNet-34~\cite{he2016deep} & 0,9 & 0,5 &  8 \\
		ResNet-50~\cite{he2016deep} & 0,5 & 0,2 &  12 \\
		MobileNet~\cite{howard2017mobilenets} & 1,8 & 1,6 &  7 \\
		MobileNetv2~\cite{sandler2018mobilenetv2} & 1,7 & 1,6 &  7 \\
		\bottomrule
	\end{tabular}
\end{table}

\begin{table}[htb]
	\caption{Accuracy of the proposed tool for different distances to the camera. The results are the average accuracy over three subjects on a TV screen with a diagonal of 108cm.}
	\label{tbl:evalACC}
	\centering
	\begin{tabular}{cccccc}
		\toprule
		1m & 2m & 3m & 4m & 5m  & 6m  \\
		\midrule
		4cm & 5cm & 8cm & 12cm & 15cm  & 19cm \\
		\bottomrule
	\end{tabular}
\end{table}

\begin{figure*}
	\centering
	\includegraphics[width=0.4\textwidth]{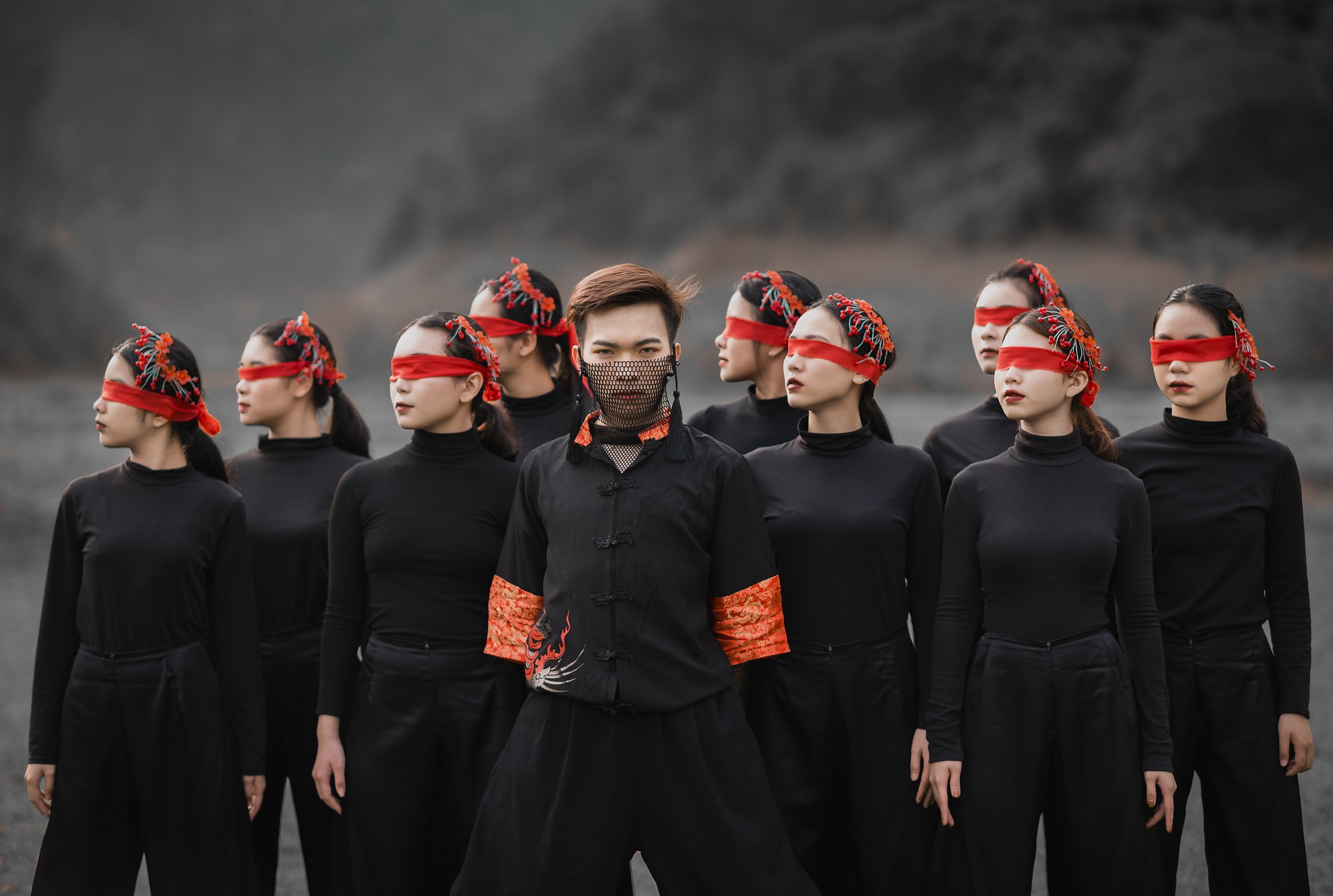}
	\includegraphics[width=0.4\textwidth]{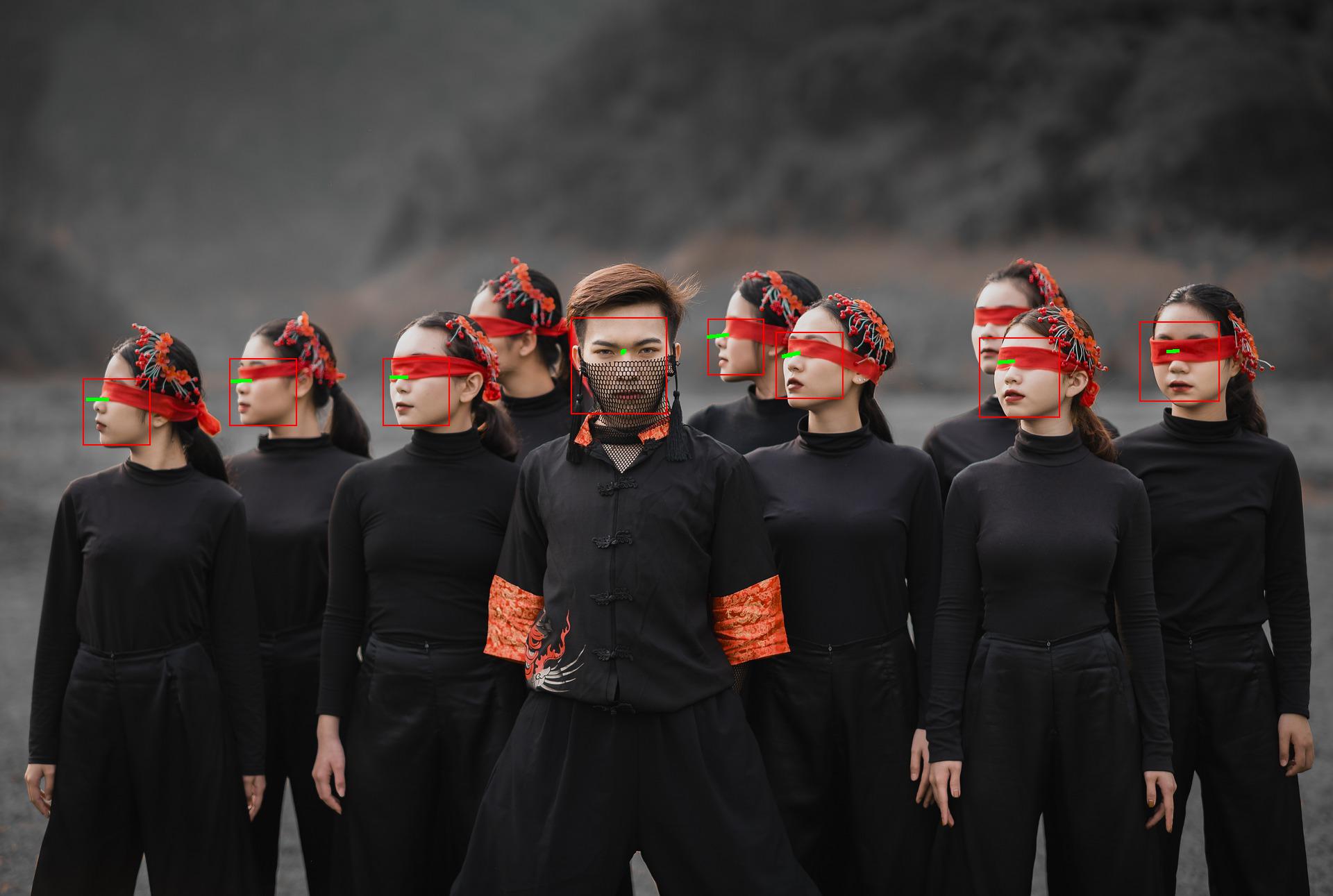}
	\includegraphics[width=0.4\textwidth]{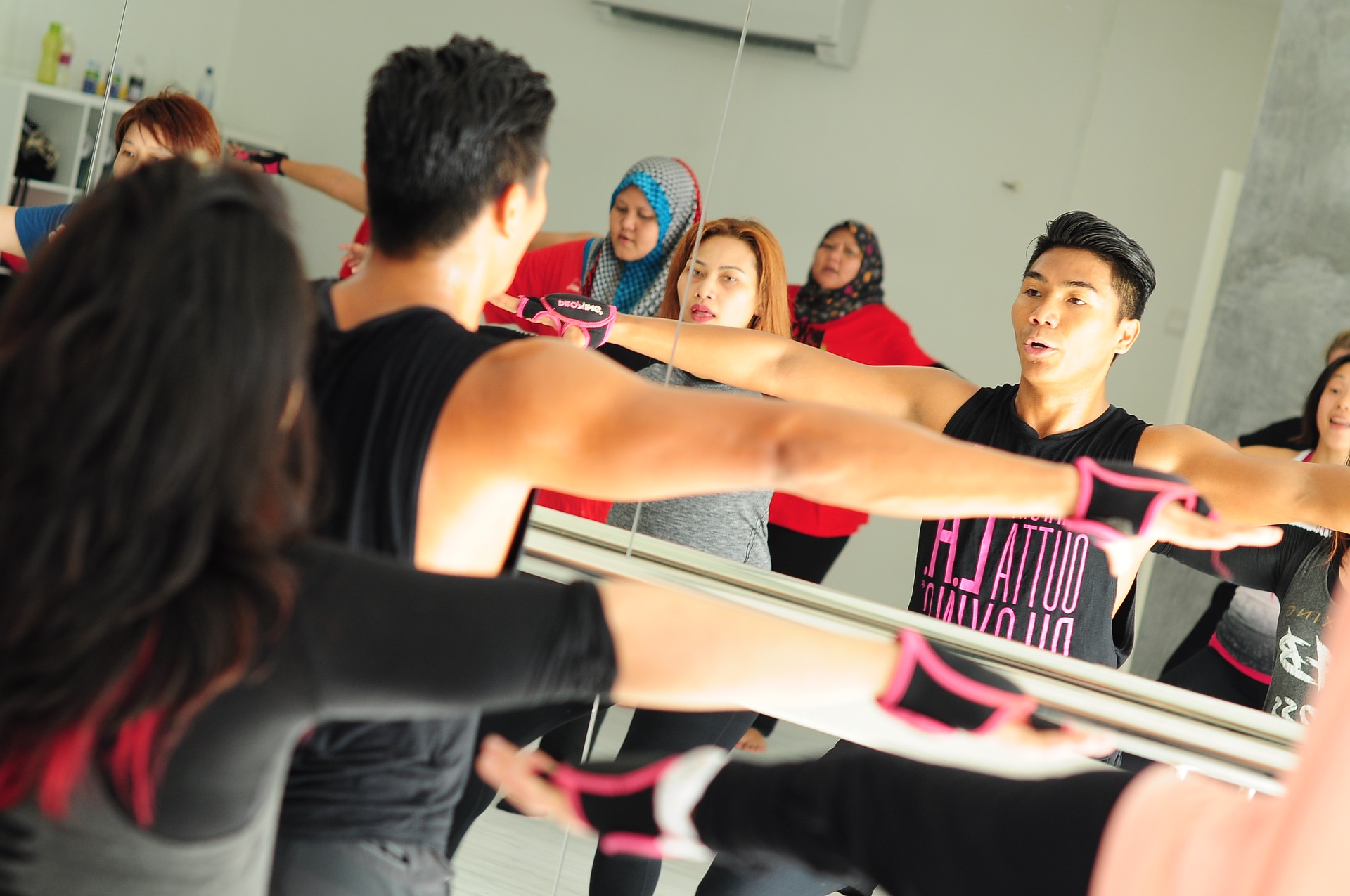}
	\includegraphics[width=0.4\textwidth]{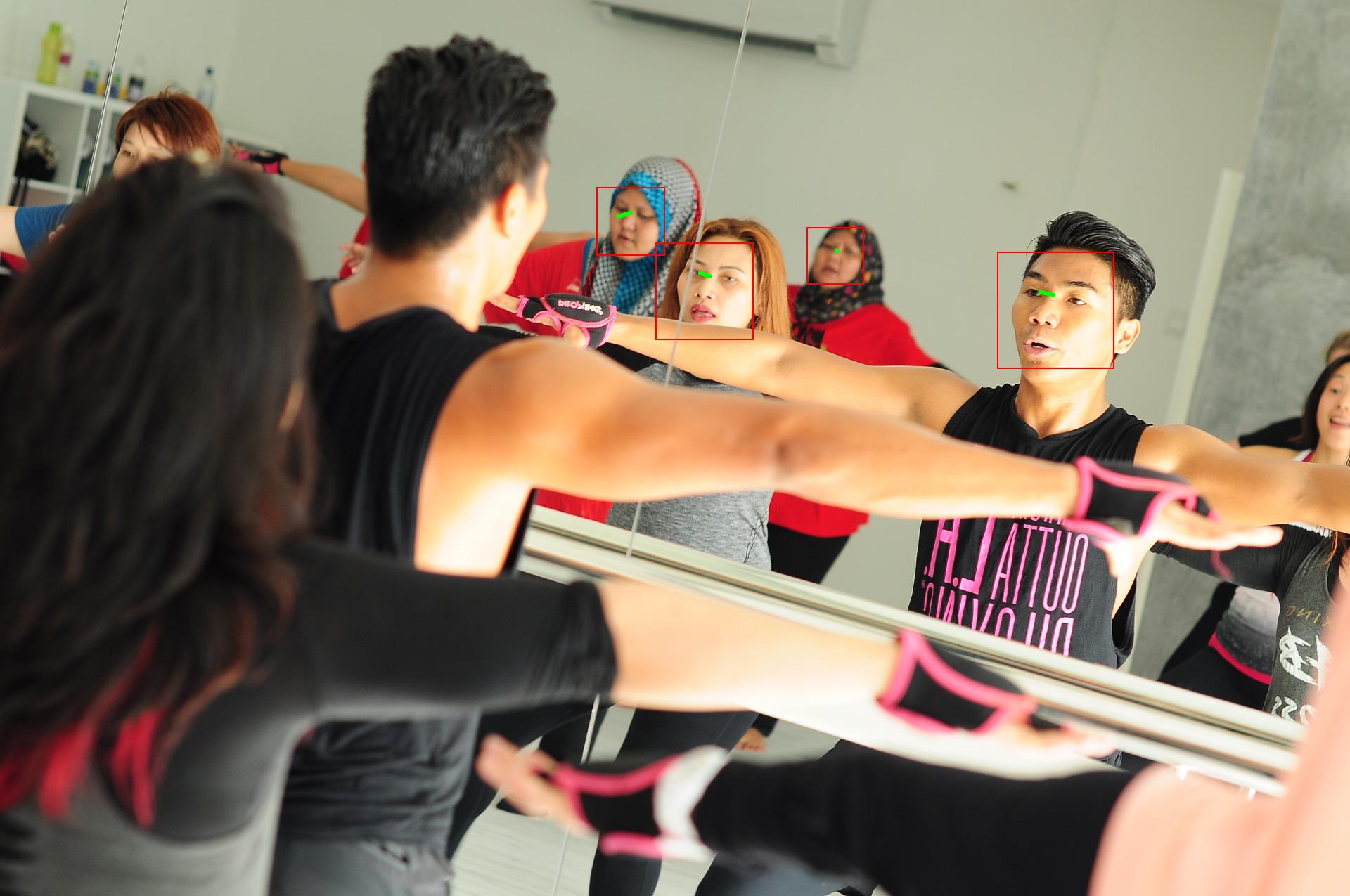}
	\includegraphics[width=0.4\textwidth]{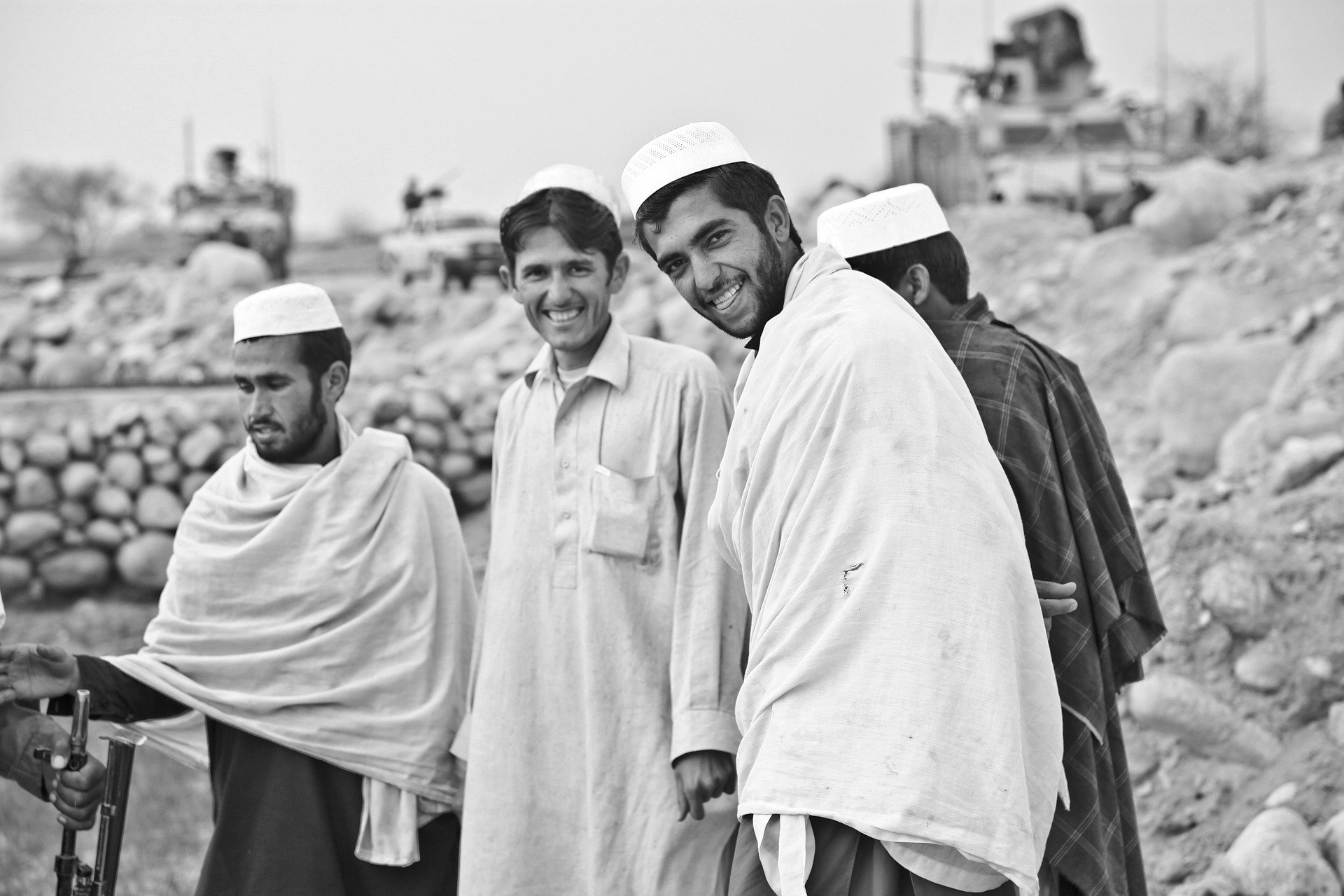}
	\includegraphics[width=0.4\textwidth]{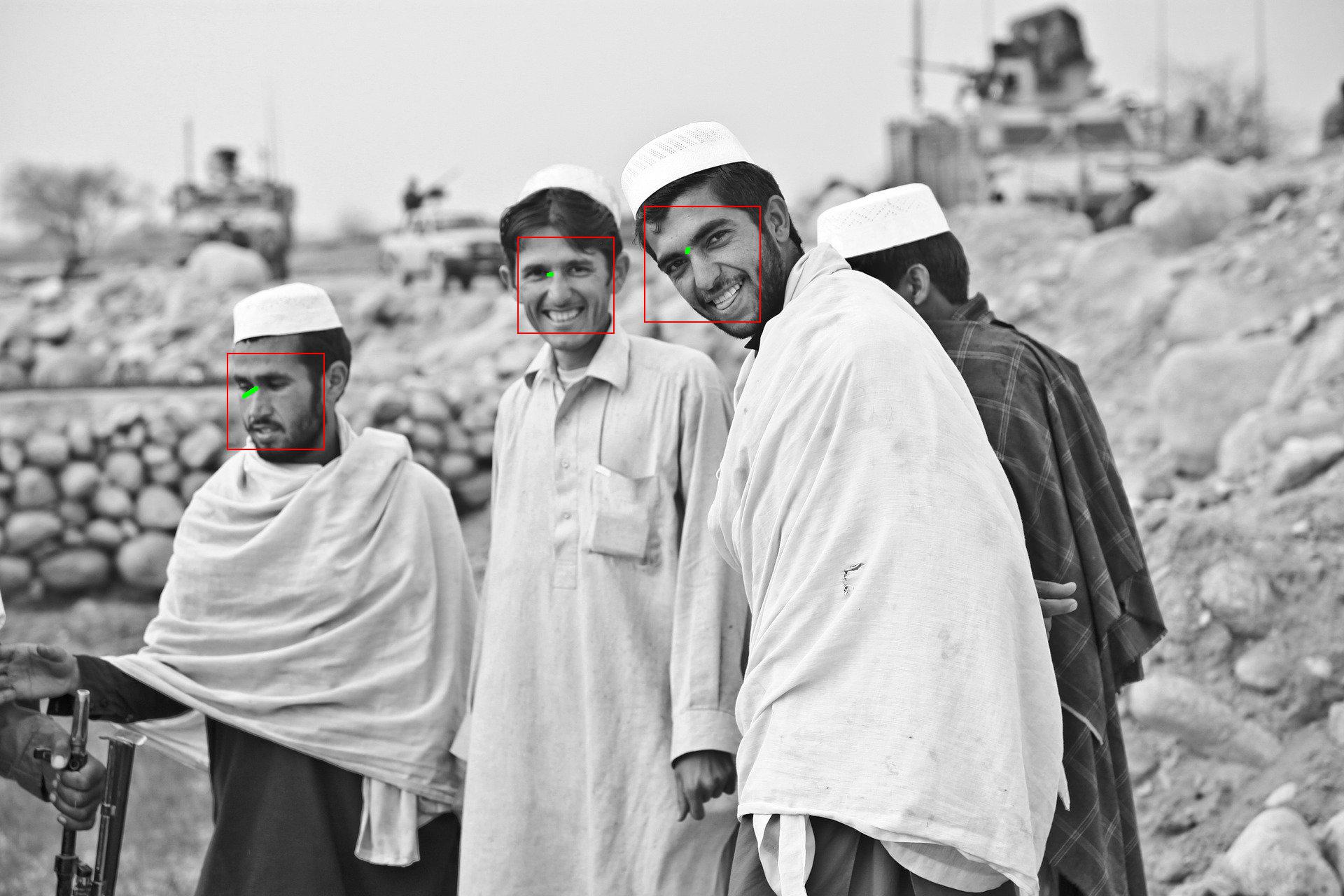}
	\includegraphics[width=0.4\textwidth]{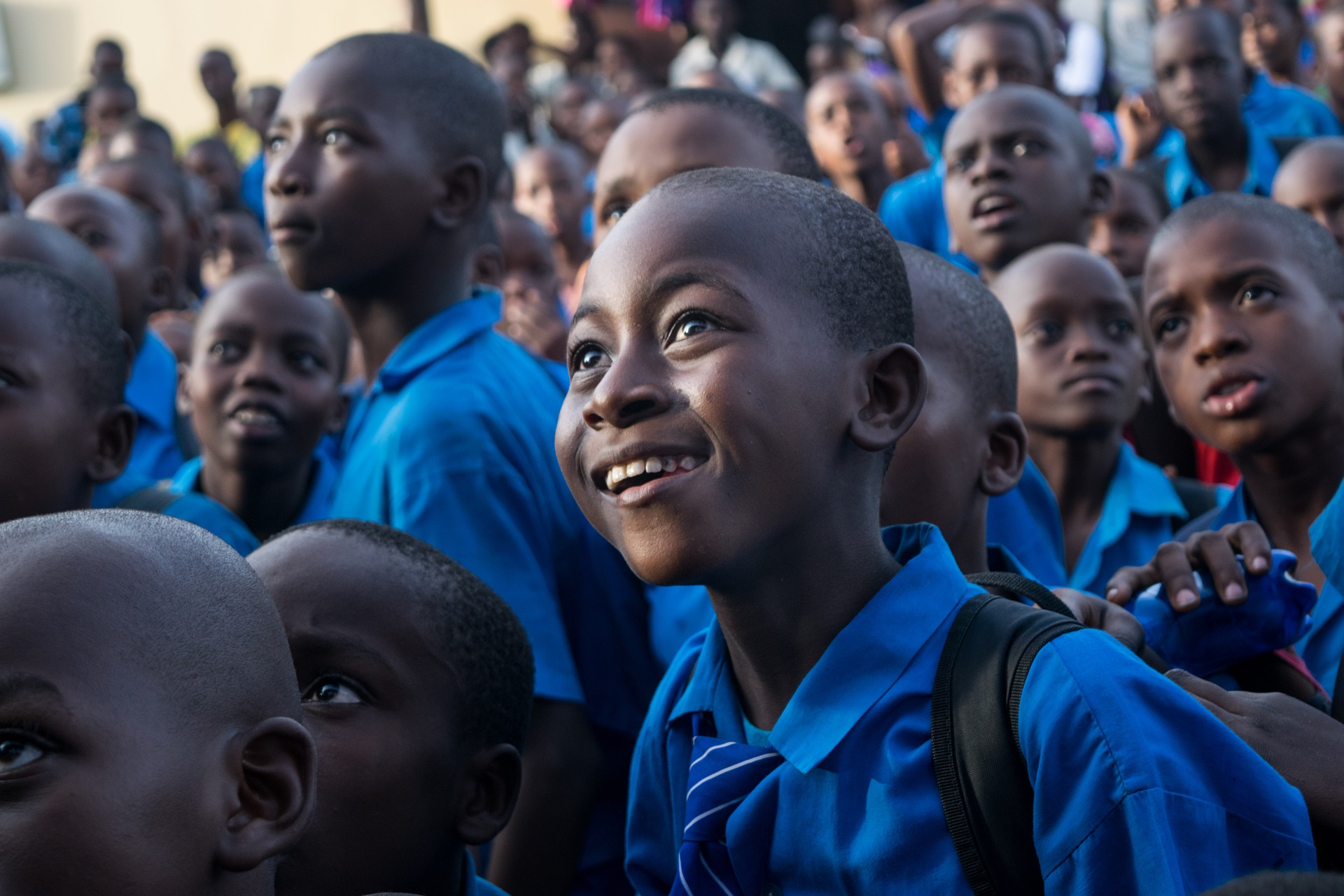}
	\includegraphics[width=0.4\textwidth]{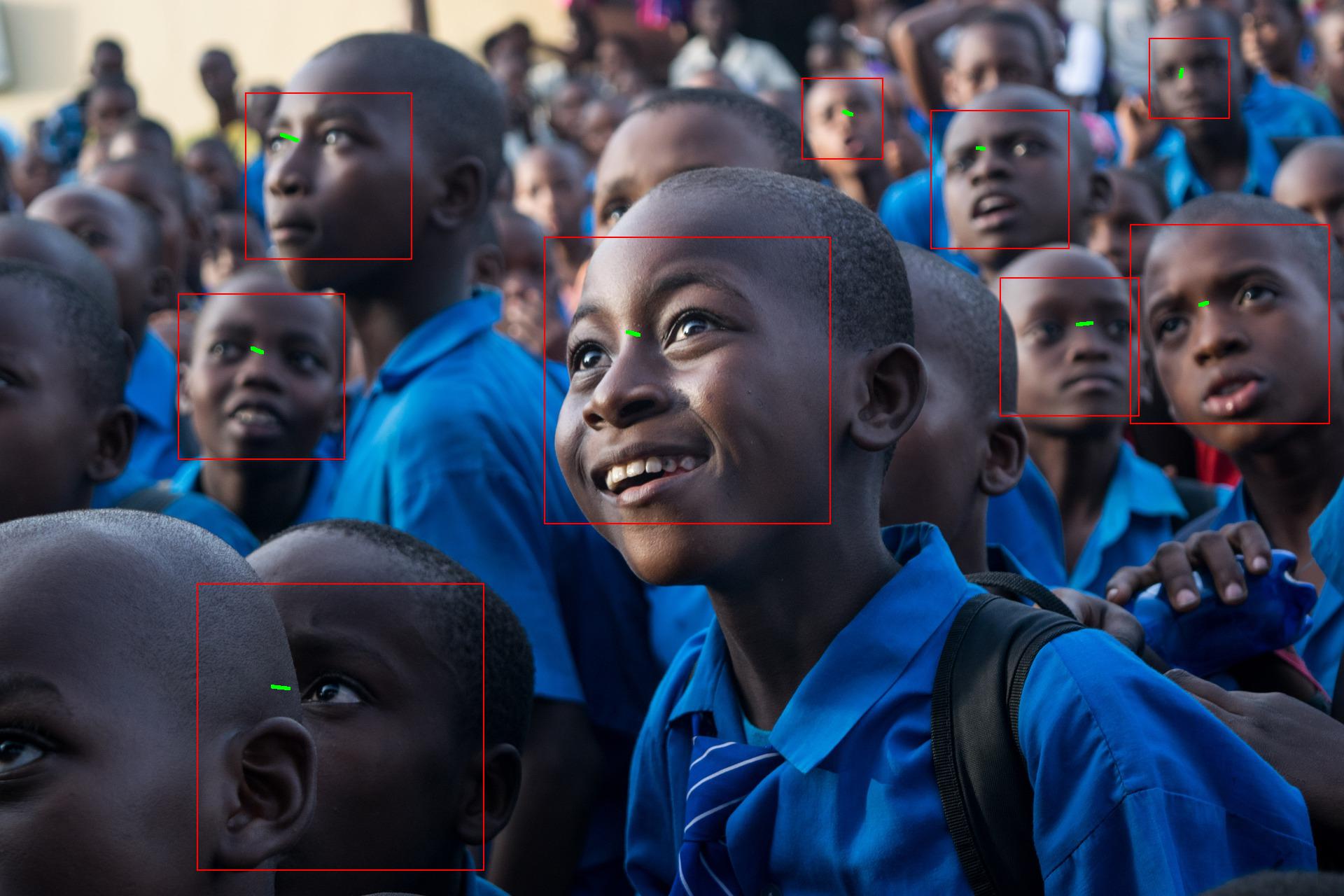}
	\includegraphics[width=0.4\textwidth]{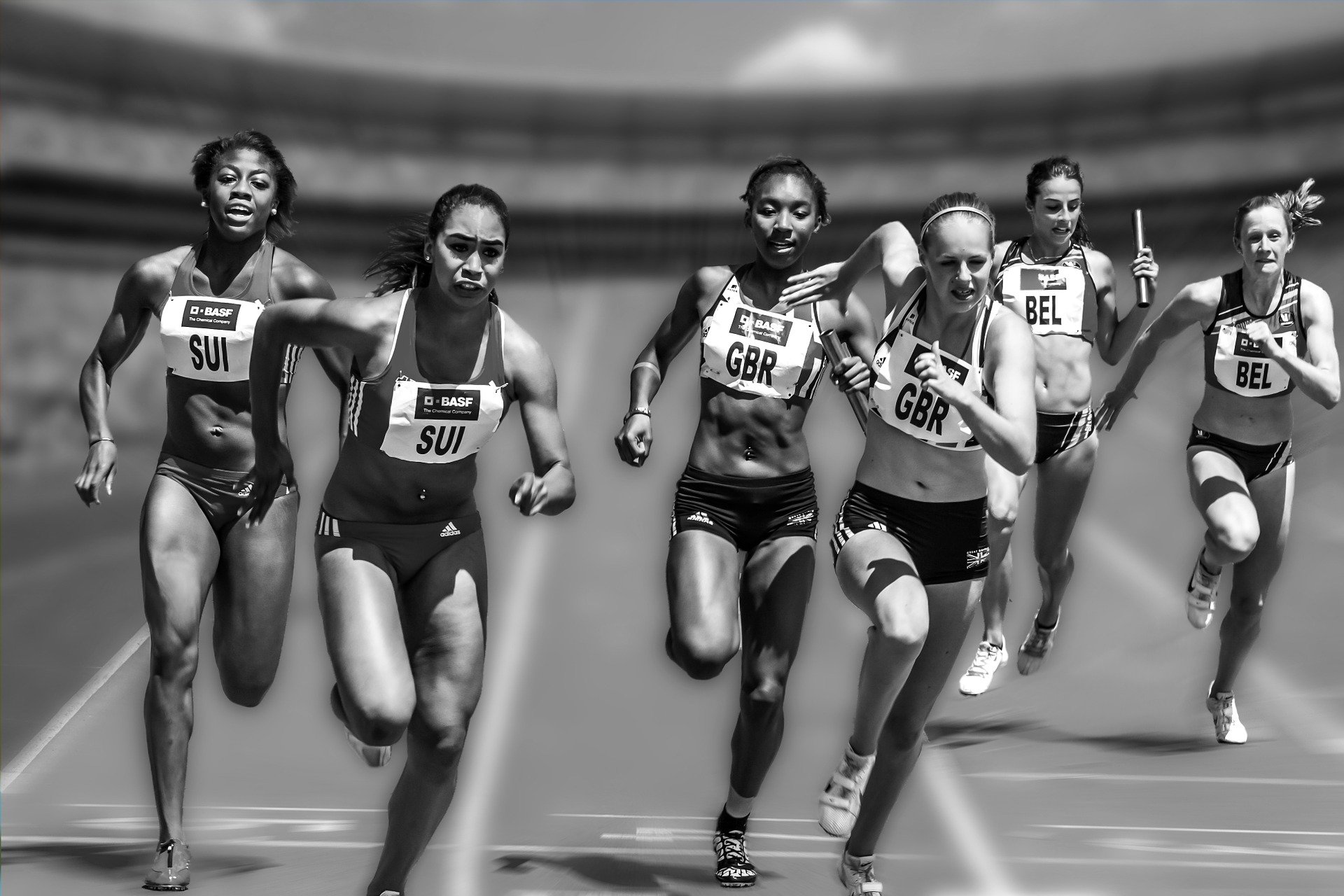}
	\includegraphics[width=0.4\textwidth]{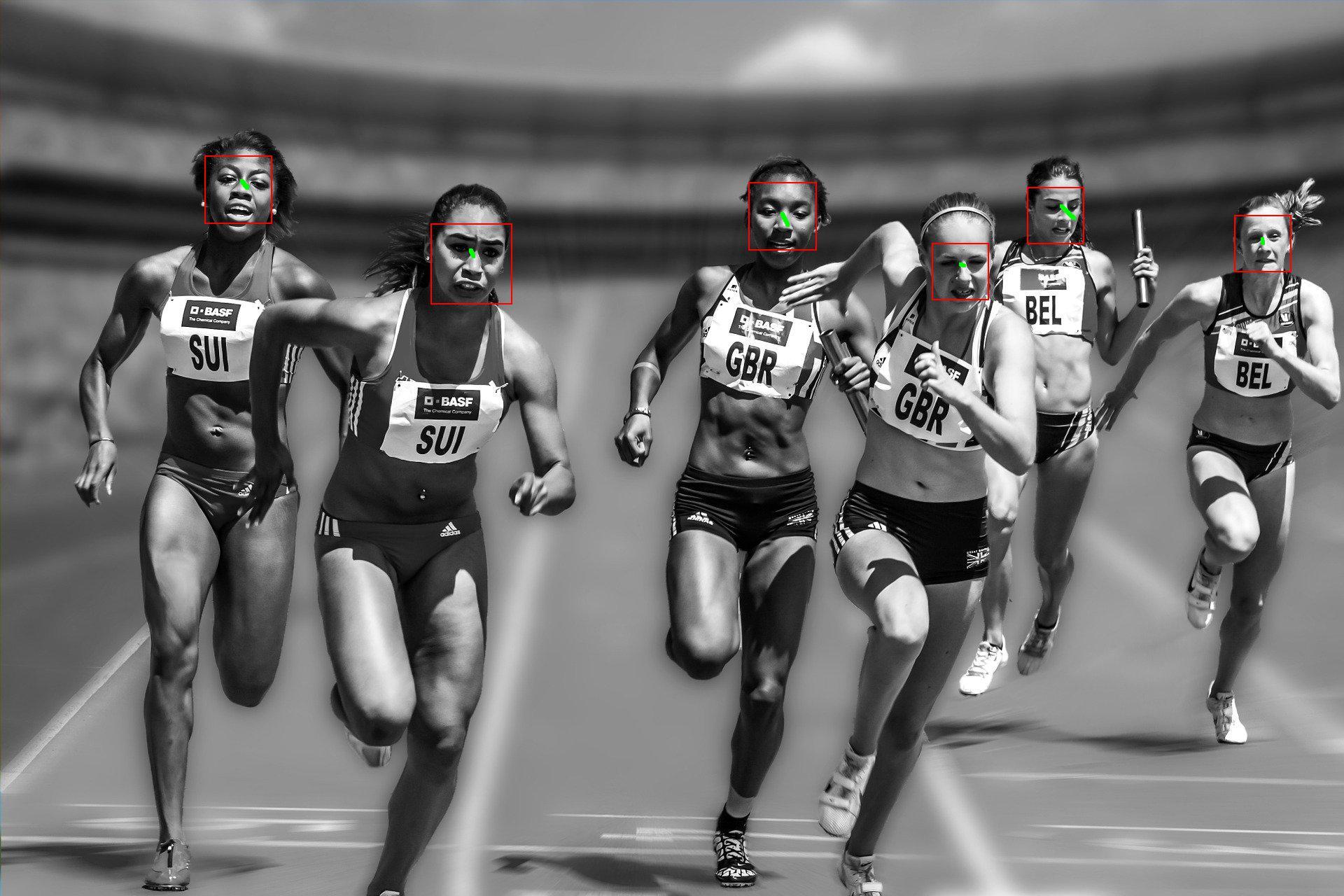}
	\caption{Qualitative evaluation on different images from pixabay.com, which are free to use. Left original image, right detections. \textbf{The images have a high resolution in the pdf so you can zoom in to see everything.}}
	\label{fig:qualieval}
\end{figure*}

In Table~\ref{tbl:evalFDDB} and \ref{tbl:evalGZ} our models are compared with other approaches. For face detection (Table~\ref{tbl:evalFDDB}), it can be seen that we have chosen a tradeoff between detection rate and runtime. The recognition rate of our approach can be further increased via the upscaling factor. However, this also increases the computation time, which also increases the runtime per image. For Yolo this is not possible, because the memory usage for images larger than $300 \time 300$ becomes too large. For the backbone of the faster-RCNN, the memory consumption is also too high for a large resolution. Which is also the main reason why we decided against YOLO and the faster-RCNN. In addition, both the faster-RCNN with backbone and the YOLO need a fixed input resolution with which they have to be trained. For our fully convolutional approach inspired by the dlib architecture, this is not necessary.

For the gaze direction determination, you can clearly see that our net runs significantly faster than the other nets. This is due to the fact that our layers use less depth than, for example, ResNet-34. The MobileNets cannot show their advantage on the GPU, since they cause cache conflicts here, whereby parts of the code are executed serialized. On a CPU, MobileNet would be significantly faster than our net, but with about 160 ms per face too slow for a real-time evaluation. In terms of results, ResNet-50 is the most accurate, closely followed by our network. In addition to accuracy, if we consider runtime on a GPU, our network is clearly ahead, which is why we chose our architecture.

The accuracy of our approach for different distances can be seen in Table~\ref{tbl:evalACC}. As can be seen, the distance has a huge impact on the accuracy. For a 6-meter distance, the error of 19 cm is approximately 20\% of the projection area. For a beamer projection this would not be as crucial since the diagonal here is usually 2 to 3 meters or more. Therefore, our system is capable of computing useful gaze positions, but does not have the accuracy of a professional remote eye tracker which can only be used from persons in front of the screen. On the other side, our system can compute and map the gaze of multiple persons in parallel and is therefore made for classrooms or meetings.

In addition to the quantitative analysis from Table ~\ref{tbl:evalFDDB} and \ref{tbl:evalGZ}, we also made a qualitative analysis. The images in Figure~\ref{fig:qualieval} have a very high resolution in the PDF, so everything can be seen in detail when using the zoom function. Also, all images are in supplementary material. Looking at the first image in Figure~\ref{fig:qualieval} you can see that our mesh can also appreciate the head-only pose when the eyes are covered. On the second, third and last image, one can see that our network works with both color and gray scale images. Likewise, it is capable of good results on crowd gatherings, as can be seen in the fourth image.

\section{Limitations}
One limitation of the presented software is that persons are not recognized via a neural network, but are assigned based on their last position. This has the limitation that the persons cannot move freely, but the software is only suitable for slightly dynamic scenarios like a classroom. In the future, recognition, which is adapted online, will be integrated.  Another limitation of our software is that currently only gaze data is extracted. Since other features such as the eyelids as well as a more accurate gaze determination via the pupils are also interesting for research, these will also be integrated in future updates. The final limitation of our software is the need for a GPU. This limitation will not be changed in the near future, since the problem of face recognition and fast gaze determination can be solved by cheaper methods, but these have strong losses in detection rate and accuracy.

\section{Conclusion}
In this paper, we have presented GroupGazer. This is a software that allows to determine the gaze direction of groups per person. This gaze determination is done online on a conventional computer with an NVIDIA GPU. GroupGazer allows each person in the group to be calibrated in parallel so that the individual gaze vectors can be mapped to a projection, such as that of a projector or large monitor. The software is intended to support behavioral research and thus make it possible to easily record the gaze positions of groups. Future enhancements to the software include person recognition for dynamic processes such as sports, pose determination, emotion determination, advanced feature extraction such as eyelids, and more accurate gaze determination based on eye features such as the pupil. We hope that the software will help other researchers in their work and further advance behavioral research as well as the application area of group-based gaze determination.

	\bibliographystyle{plain}
	\bibliography{template}

\end{document}